\documentclass[acmlarge]{acmart}
\setcopyright{none}
\renewcommand\footnotetextcopyrightpermission[1]{}
\authorsaddresses{}
\acmDOI{}
\acmVolume{}
\acmNumber{}
\acmArticle{}
\acmYear{}
\acmMonth{}

\usepackage{booktabs}
\usepackage{array}
\usepackage{algorithm}
\usepackage{algorithmic}
\usepackage{etoolbox}
\graphicspath{{./}}

\setcitestyle{super,open={[},close={]}}
\begin{document}

\title{How Early Is Early Enough? Design-Dependent Observation-Window
Sufficiency in Subscription Churn Prediction}

\author{Xiao Han}
\affiliation{
  \institution{Emory University}
  \department{Goizueta Business School}
  \city{Atlanta}
  \state{GA}
  \postcode{30322}
  \country{USA}
}
\email{xhan@alumni.emory.edu}

\author{Yao Xiao}
\affiliation{
  \institution{Georgia Institute of Technology}
  \department{College of Computing}
  \city{Atlanta}
  \state{GA}
  \postcode{30332}
  \country{USA}
}
\email{yxiao344@gatech.edu}

\author{Chenyu Wu}
\affiliation{
  \institution{Duke University}
  \department{Pratt School of Engineering}
  \city{Durham}
  \state{NC}
  \postcode{27708}
  \country{USA}
}
\email{wuchenyu999@outlook.com}

\author{Tongchen Zhang}
\affiliation{
  \institution{Independent Researcher}
  \city{Seattle}
  \state{WA}
  \postcode{98101}
  \country{USA}
}
\email{tzhangaca@gmail.com}

\begin{abstract}
How many days of early behavior suffice for subscription churn prediction? In
the public KKBox dataset, the early indicator of churn is typically an indicator
of someone's contract status; however, when looking in the heavily churned
manual-renewal segment, having access to early behavior creates a substantial
increase in prediction for that specific segment (PR~$+0.10$ at 120~days). A
nine-window sufficiency curve shows diminishing returns; descriptive inflection
detectors place a knee between 30 and 120~days depending on the metric and
model class. Stress-testing over three cohort/task
designs shows that this curve is singular to the design being tested; for
example, in our test with a moving target, the curve inverts and can shift
depending on the feature set used. Therefore, any window-sufficiency claim
should state its cohort construction, target definition, and feature families.
A survival-filtered replication on educational-attrition
data (OULAD) confirms the rising-curve shape in one behavior-dominated
setting (floor ROC~0.56 vs.\ 0.88), though signal composition inverts.
\end{abstract}




\makeatletter
\renewcommand{\@mkauthors}{\ignorespaces}
\pretocmd{\@mkabstract}{%
  {\small\noindent
  \textbf{Xiao Han}$^{*}$\\
  Emory University, Goizueta Business School, Atlanta, GA, USA, xhan@alumni.emory.edu\\[0.4em]
  \textbf{Yao Xiao}\\
  Georgia Institute of Technology, College of Computing, Atlanta, GA, USA, yxiao344@gatech.edu\\[0.4em]
  \textbf{Chenyu Wu}\\
  Duke University, Pratt School of Engineering, Durham, NC, USA, wuchenyu999@outlook.com\\[0.4em]
  \textbf{Tongchen Zhang}\\
  Independent Researcher, Seattle, WA, USA, tzhangaca@gmail.com\\[0.2em]
  {\scriptsize $^{*}$Corresponding author.}\par}%
  \vspace{0.3em}%
}{}{}
\makeatother
\maketitle
\thispagestyle{empty}
\fancyhf{}\renewcommand{\headrulewidth}{0pt}\renewcommand{\footrulewidth}{0pt}
\cfoot{\thepage}
\pagestyle{fancy}

\section{Introduction}
Subscription-based companies should choose when to intervene to keep a customer;
if they intervene late, the customer may be gone and no marketing tools will be
useful; if they intervene early when they have limited data, the interventions
are likely to be wrong. A good way to explore this question is to create a
sufficiency curve: chart performance by the number of days of early activity the
model may observe, find the point where diminishing returns start occurring, and
locate the shortest sample period which yields the greatest part of the
attainable signal. The goal is a two-pronged answer: an operational number
(``$N$ days is sufficient'') and, for a finance application, transparency
regarding which customer characteristics comprise the rationale.

Two separate studies established prior findings. Ballings and Van den
Poel~\cite{ballings} (``Customer event history for churn prediction: How long is
long enough?'') varied the number of years of existing customer history and
found diminishing returns in a log function; nearly all historical depth could
be discarded with little loss of AUC, replicated in a retail
setting~\cite{apaydin}. Drachen et~al.~\cite{drachen} noted that players' churn
can be partially predicted at early points in their engagement (first
session/day/week), but with three coarse points, accuracy only, and no formally
located inflection. Because these studies fixed a specific cohort design and
measured only a single curve, neither can test whether the inflection remains
valid after varying any attributes of that design.

Using the public KKBox dataset, we constructed an early-observation-window
sufficiency curve for subscription churn, identified the early warning signal,
then used three designs to stress-test the curve by progressively controlling
\emph{what} changes as the window grows.

\textbf{Primary contribution: window-sufficiency results depend on
experimental design, not just data.} A nine-window sufficiency curve on KKBox,
stress-tested across three cohort/task designs, yields a rising curve (fixed
horizon), a declining/U-shaped curve (moving-target landmark), and different
knee locations depending on the feature set. A survival-filtered replication on
educational-attrition data (OULAD) confirms the rising-curve shape in one
behavior-dominated setting whose signal composition inverts KKBox's.
Two supporting findings:
\begin{itemize}
\item \textbf{Contract-driven early signal with behavioral lift in churn-dense
segments.} The contract floor dominates overall, but on the manual-renewal
segment (28\% of users, 86.6\% of churners) the floor collapses to ROC~0.63
and behavior adds real lift (PR~$+0.10$ by day~120).
\item \textbf{Early predictability survives without survivorship conditioning.}
We ran a separate unconditioned control (Design~U) on all 1.97M acquirees
with no survival filter. It already hits ROC~0.913 at day~7. Plan duration
does partly encode the decision horizon, so this is an upper bound.
\end{itemize}

\section{Related Work}
Ballings and Van den Poel~\cite{ballings} asked ``how long is long enough?''
and varied years of retained customer history, finding logarithmic diminishing
returns (replicated in grocery by~\cite{apaydin}). But their axis is the
depth of an \emph{existing} customer's record, not early post-acquisition
behavior. They also report only one curve under one design, which is exactly
what this paper stress-tests. Drachen
et~al.~\cite{drachen} demonstrated an early observation window for game
retention (the only prior study varying an \emph{early} window) but did not
formally establish a knee; they had three data points for accuracy only.
Game-churn studies~\cite{runge,hadiji} and Bhattacharjee
et~al.~\cite{bhatt} fix a single window and do not vary the cohort
\emph{design}. Our Designs~B/C are \textbf{landmark} models as defined by van
Houwelingen~\cite{vh07}: at each landmark $N$ we predict the future event from
subjects still at risk, using covariates observed up to~$N$;
Sandqvist~\cite{sandqvist2026} recently extended landmarking supermodels
with gradient boosted trees for time-dependent covariates. The
competing-risks framework~\cite{finegray} provides rationale for treating
early-leavers as a resolved outcome rather than a dropped sample.
Kimitei et~al.~\cite{kimitei2025} demonstrated the predictive and
explanatory value of survival models for churn. Landmarking is
a well-established technique; we use it to explain why a sufficiency curve takes
its shape. The WSDM Cup 2018 KKBox challenge~\cite{wsdm} defines churn as no
renewal within 30~days of expiration; we re-derive the label from transactions
to avoid the known ${\sim}1\%$ expiry-date leak (\S\ref{sec:method}). To our
knowledge this is the first early-observation-window sufficiency curve on a
large public subscription dataset; supporting early-signal~\cite{khan} and
survey~\cite{review} work notes the time-window dimension is under-studied.
Mufti et~al.~\cite{mufti2026} recently proposed a rolling-window framework
for non-contractual churn with explicit temporal separation between
observation and evaluation windows.
Joy et~al.~\cite{joy2024} combined recurrent temporal layers with
LightGBM for streaming-service churn, achieving AUC~0.956 and
confirming gradient boosting's competitiveness in subscription
retention tasks.
The broader question of when additional data ceases to improve model
performance arises across domains, including model
compression~\cite{li2024sec}.

\section{Method}\label{sec:method}
\textbf{Dataset and label.} KKBox (WSDM Cup 2018): \texttt{members},
\texttt{transactions} (plan/price/auto-renew/cancel/expiry), and ${\sim}30$\,GB
of daily \texttt{user\_logs}. We \textbf{re-derive} the churn label from
\texttt{transactions} following the official logic (no valid renewal within
30~days of expiration), anchoring windows cleanly and avoiding the known expiry
leak. The re-derived labels agree with \texttt{train\_v2} on \textbf{92.1\%} of
overlapping users (anchor-driven gap, not label noise; mechanism match:
95.7\%); refitting with the official label reproduces the rising curve.

\textbf{The three designs (the worked stress-test).} Three anchorings are used
to study the same day-since-acquisition sweep. $t_0 =$ first transaction date
throughout; the feature horizon $[0,N)$ grows with $N$ in all three. These are
diagnostic designs, not three competing models.

\emph{Design A (primary).} Label fixed at $t_0{+}180$; cohort: acquirees with
churn-defining expiry $\geq t_0{+}180$, resolvable label, and $\geq$1 in-window
active log day. This is a \emph{survival-conditioned} cohort (``for a subscriber
who reaches the $\geq$180-day renewal decision, how early is it
predictable?''). The $\geq$180 filter leaves \textbf{1{,}294{,}009 of
2{,}426{,}143} acquirees (dropping \textbf{47\%}); resolvable label + $\geq$1
active day leaves the \textbf{1{,}040{,}946} modeled cohort, churn
\textbf{8.37\%}. Features: 48 columns (6 member $+$ 11 tx $+$ 31 log).

\emph{Design B (all-acquirees landmark).} No survival filter:
\textbf{1{,}973{,}026 (81.3\%)} acquirees, prevalence 0.4264. At landmark $N$,
keep users still subscribed and predict churn in the residual $(N,180]$, the
design an operator could deploy at day~$N$. \emph{Three} things move with $N$:
feature horizon grows, residual target $(N,180]$ shrinks, and cohort re-selects
to survivors-to-$N$. As $N$ increases, the prediction window shrinks and the
remaining cohort shifts toward healthier survivors; these target and cohort
changes can cause performance to \emph{decline} despite longer observation. Features: 17
member$+$tx only (no log time series, so the CNN cannot run on this design).

\emph{Design C (controlled landmark).} To ask ``does the rise survive when
\textbf{only} the feature horizon moves?'', we fix cohort and target: cohort $=$
resolvable acquirees still subscribed at a single landmark $L{=}120$
($n{=}$\textbf{1{,}263{,}885}, identical for every $N$); target $=$ churn in
$(120,180]$ (prevalence 0.1358, constant). Only $[0,N)$ varies. This removes
B's moving target and re-selection, relaxes A's $\geq$180 filter to a milder
$\geq$120, but \emph{does not} remove survival conditioning (\S\ref{sec:disc}).
Features: same 17 as B (so A vs.\ C also differ in feature set).
Table~\ref{tab:designs} summarizes the three designs.

\begin{table}[t]
\caption{Experimental designs. All share the same growing feature window
$[0,N)$; they differ in what else moves with $N$. A separate Design~U
(unconditioned control, \S\ref{sec:findings}) shares B's 1.97M cohort but fixes
the target at churn-by-180 with no survival filter.}
\label{tab:designs}
\centering
\small
\setlength{\tabcolsep}{4pt}
\begin{tabular}{@{}lp{1.6cm}p{1.8cm}p{1.6cm}@{}}
\toprule
 & \textbf{A} (primary) & \textbf{B} (landmark) & \textbf{C} (control) \\
\midrule
Cohort & Fixed: surv.\,$\ge$180\,d & Changes: alive at $N$ & Fixed: surv.\,$\ge$120\,d \\[3pt]
Target & Fixed: churn at $t_0{+}180$ & Moves: churn in $(N,180]$ & Fixed: churn in $(120,180]$ \\[3pt]
Factors & \textbf{1} & \textbf{3} & \textbf{1} \\
\bottomrule
\end{tabular}
\end{table}

\textbf{Leakage controls.} Features use only log rows with
\texttt{day\_offset}$<N$ and transactions with $0\le\mathtt{tx\_date}-t_0<N$
(strict); Design~A's label is fixed at the $t_0{+}180$ cut; in B/C landmark
status uses only $\mathtt{tx\_date}\le t_0{+}N$. We do not use the leaked
expiry file, and per-window assertions plus tabular$\leftrightarrow$sequence
cross-checks pass.

\textbf{Features and models.} The feature set is held \emph{identical across
all $N$} so the curve isolates the window effect: in-window \texttt{user\_logs}
aggregates (listen volume, active-day count/rate, completion-mix shares, daily
mean/variance, inter-active-day gaps, OLS trend slopes), \texttt{members}
(city, age, gender, registered-via), and early \texttt{transactions} (plan,
price, auto-renew, cancellation). Two representations come from the same daily
rollup: a tabular matrix (GBDT) and a length-180 daily sequence (window $N=$
prefix slice) for the CNN. A \emph{required} $N{=}0$ static-only GBDT (17
member$+$tx columns) separates ``value of a profile'' from ``value of
behavioral observation.'' For robustness we also include a \textbf{matched-size
GBDT} trained on the identical 250k-row subsample the CNN uses, and
single-feature rankers (recency-only, price-only). We sweep \textbf{nine
windows ($N{=}7\ldots180$) plus $N{=}0$} in Design~A and eight
($N{=}7\ldots120$) in B/C. A and C share one stratified 5-fold split (seed 42)
reused for every (window$\times$model) so comparisons are paired; B builds folds
per-$N$.

\textbf{Metrics.} PR-AUC and ROC-AUC (threshold-free primaries). We
report \textbf{5-fold means with fold-to-fold spread}, not paired-$t$ CIs
(folds share ${\sim}80\%$ data, so $t(4)$ is anti-conservative
\cite{dietterich,nadeau}). Fold spread is small (Design~A ROC $\pm0.002$), so
we lead with effect sizes and non-overlapping fold ranges. Inflection detection:
Kneedle~\cite{kneedle} as a \emph{descriptive locator} plus a marginal-gain
threshold; within-method stability is assessed by a 1{,}000-replicate
user-level bootstrap (resample users with replacement from out-of-fold
predictions, recompute AUC per window, re-run Kneedle per replicate).
Across designs we compare \textbf{shape}, not level, since cohorts
differ in prevalence and feature set.

All models are frozen across windows. The \emph{primary} GBDT is LightGBM~\cite{lgbm}
(lr\,$=$\,0.05, num\_leaves\,$=$\,63, min\_child\,$=$\,200,
feature/bagging\_frac\,$=$\,0.8, 3000 rounds, early-stop 100 on val AUC;
categoricals: city, reg\_via, gender).
A 1D-CNN (2$\times$Conv1d(32,$k{=}3$)$+$BN$+$ReLU, global pool; 250k rows,
patience~2) and logistic regression (L2, $C{=}1$) round out model-class
robustness. Design~A uses 48 features (6 member $+$ 11 tx $+$ 31 log); B and C
use 17 member$+$tx only (no logs).

\begin{algorithm}[t]
\setlength{\abovecaptionskip}{2pt}
\setlength{\belowcaptionskip}{-2pt}
\caption{Window-Sufficiency Evaluation}\label{alg:sufficiency}
\footnotesize
\setlength{\abovedisplayskip}{0pt}
\setlength{\belowdisplayskip}{0pt}
\begin{algorithmic}[1]
\REQUIRE Dataset $\mathcal{D}$, windows $\mathcal{N}{=}\{0,7,\ldots,120\}$, design $d{\in}\{A,B,C,U\}$, $K$ folds
\ENSURE Sufficiency curve $\{(N, \text{AUC}_N)\}$, inflection band
\FOR{fold $k = 1 \ldots K$}
  \FOR{window $N \in \mathcal{N}$}
    \STATE Build cohort/target per $d$; extract features from $[0,N)$
    \STATE Train model on fold-$k$ train; evaluate AUC on fold-$k$ test
  \ENDFOR
\ENDFOR
\STATE Mean AUC curve $\to$ Kneedle knee $\hat{N}$; verify non-overlapping fold ranges
\STATE Bootstrap ($B{=}1000$): resample users, recompute AUC \& Kneedle $\to$ 95\% CI
\STATE Cross-method comparison $\to$ \textbf{descriptive inflection band}
\end{algorithmic}
\vspace{-4pt}
\end{algorithm}
\vspace{-2pt}

\textbf{Cross-domain dataset for replication.} In order to investigate
if the phenomenon of rising curves is peculiar to subscription models,
we test the replication on the Open University Learning Analytics Dataset
(OULAD~\cite{oulad}): 32{,}314 student-module enrollments (7 modules,
4~semesters). Online learning platforms collect massive behavioral log
data that can contribute to the evaluation of instructional
technologies~\cite{li2023edu} and also allow predicting outcomes. Since
VLE activity ceases at withdrawal, students who withdrew within the
observation period are excluded since only past events can be detected.
We therefore employ a Design-C construction: cohort $=$ the 24{,}484
enrollments still registered at landmark $L{=}120$ (withdrawal rate
9.1\%); target $=$ withdraw after day~120; features grow within $[0,N)$,
$N{=}7{\ldots}120$. Static features: 7 demographics. Behavioral
features: VLE click aggregates plus assessment submissions, growing
with~$N$. Same LightGBM setup, 5-fold stratified CV.

\section{Results}

\subsection{The fixed-horizon curve rises; the knee is method-dependent}
The data from Table~\ref{tab:curve} indicates that Design~A's GBDT
curve moves upwards consistently, albeit showing signs of diminishing
returns. From $N{=}7$ to $N{=}120$, the PR figure shows an effect of
\textbf{$+0.069$}, and the ROC obtains \textbf{$+0.021$}, along with
tight and \textbf{non-overlapping fold ranges}, indicating the upward
movement is not fold noise. When compared to the early-segment peak
attainability at $N{=}120$, \textbf{62\% of behavioral PR-lift by
day~120 was observed by day~45, while 72\% was reported by day~60, and
90\% by day~90.}

\begin{table}[t]
\caption{Design-A GBDT sufficiency curve (mean over 5 folds; fold spread ROC
$\le\pm0.002$, PR $\le\pm0.006$). $N{=}180$ is a separate near-decision regime
(\S\ref{sec:tworegime}); every headline number is from $N\le120$.}
\label{tab:curve}
\centering
\small
\begin{tabular}{rccc}
\toprule
$N$ (days) & PR-AUC & ROC-AUC & \% of PR gain \\
\midrule
0 (floor) & 0.367 & 0.878 & 0\% \\
7   & 0.387 & 0.880 & 23\% \\
14  & 0.397 & 0.882 & 33\% \\
21  & 0.405 & 0.884 & 42\% \\
30  & 0.411 & 0.887 & 49\% \\
45  & 0.422 & 0.891 & \textbf{62\%} \\
60  & 0.431 & 0.894 & \textbf{72\%} \\
90  & 0.447 & 0.899 & \textbf{90\%} \\
120 & 0.456 & 0.901 & 100\% \\
\textit{180 (near-dec.)} & \textit{0.520} & \textit{0.919} & \textit{---} \\
\bottomrule
\multicolumn{4}{@{}p{0.95\columnwidth}@{}}{\scriptsize%
Non-overlapping check: at worst-case spread ($\pm0.002$ ROC, $\pm0.006$ PR),
$N{=}7$ ROC $\in[0.878,0.882]$ and $N{=}120$ ROC $\in[0.899,0.903]$;
$N{=}7$ PR $\in[0.381,0.393]$ and $N{=}120$ PR $\in[0.450,0.462]$.}
\end{tabular}
\end{table}

Although the location of a knee cannot be reliably established given the
disparities among the detectors, Kneedle (a \emph{descriptive locator},
not a formal statistical test) typically refers the knee to 45--60~days,
while the marginal-gain threshold spans 30--120~days across models.
Logistic regression's curve is too gradual to identify any formal knee.
We summarize this as a \textbf{descriptive inflection band}. A
1{,}000-replicate user-level bootstrap of the Kneedle detector yields a
95\% CI of $[45,60]$ on both ROC and PR (ROC knee at 45~days in 69.6\%
of replicates, 60~days in 30.4\%; PR: 51.4\%/48.6\%; 4 PR replicates
found no concave knee). The uncertainty within each method remains
narrow; it is the spread between methods (Kneedle 45--60 vs.\
marginal-gain 30--120 vs.\ LogReg: none) that dominates, confirming the
band width is method-driven, not data-unstable. On a log-day axis the
curve is near-linear ($R^2 \approx 0.96$ for both metrics), so the
concavity Kneedle detects is partly an axis-parameterization artifact,
which is itself another dimension of design-dependence.

\textbf{Two regimes.}\label{sec:tworegime} The $120{\to}180$ step
re-accelerates (PR~$+0.064$, ROC~$+0.018$): \textbf{endpoint proximity}, not
added history (84.1\% of users decide within 30~days of $N{=}180$;
${\approx}\tfrac{2}{3}$ of the jump is proximity). \textbf{Every headline
number is from $N\le120$.}

\subsection{A worked stress-test: where the rise, shape, and knee come from}\label{sec:stresstest}
\begin{figure*}[t]
\centering
\includegraphics[width=0.98\textwidth]{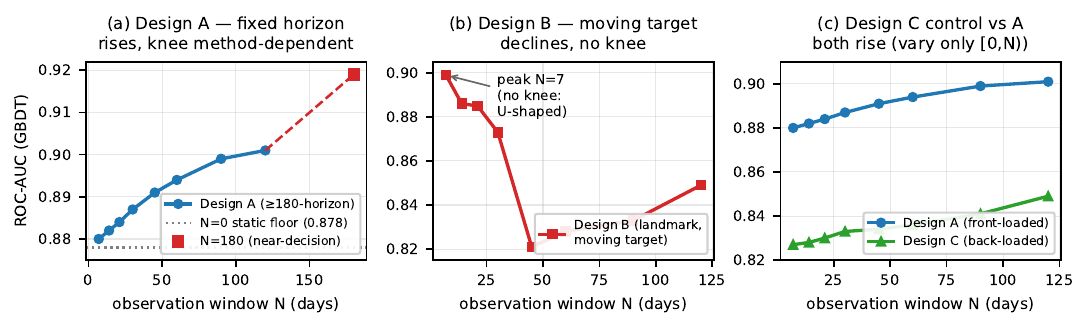}
\caption{Three-design stress-test (GBDT ROC-AUC vs.\ window; KKBox).
(a)~A: rises, descriptive knee method-dependent; $N{=}180$ is near-decision.
(b)~B: declines/U, no knee.
(c)~C and A both rise; C is back-loaded.}
\label{fig:ladder}
\end{figure*}

Using the three-design ladder (Figure~\ref{fig:ladder}), we identify which
portions of the curve are design-dependent.

\textbf{(1) The rise is not an artifact of the moving target or per-$N$
re-selection.} Design~C tests this directly: fixing the cohort and target,
removing Design~B's moving target and re-selection and relaxing Design~A's
filter to a \emph{milder, fixed $\geq$120}, the curve still \textbf{rises
monotonically}, with total gain \textbf{PR $+0.0656$ / ROC $+0.0213$} over
$[7,120]$, essentially identical to Design~A's $+0.0687$ / $+0.0210$, with
non-overlapping endpoint fold ranges (fold spread ROC $\le\pm0.003$, PR
$\le\pm0.003$). The unconditioned Design~U (\S\ref{sec:findings}, no survival filter)
provides independent confirmation. The informative result is the
\emph{magnitude match} to Design~A and the front-vs-back-loading contrast
in~(3).

\textbf{(2) The \emph{shape} depends on the prediction task: two legitimate
questions, not a paradox.} The shape of a sufficiency curve is contingent
upon the prediction task; there are two correct questions being asked,
not a paradox. When we re-pose the sweep as a moving-target landmark
(Design~B), an inversion of the curve occurs: GBDT ROC achieves its
maximum at $N{=}7$ (0.899), falls to 0.821 at $N{=}45$, and recovers to
0.849 only at $N{=}120$. The question is why: as $N$ increases, the
residual window $(N,180]$ gets shorter and the cohort re-selects to
healthier survivors. Prevalence drops from $0.385$ (at $N{=}7$) to
$0.136$; this can be attributed to the covariate
shift---a distributional displacement quantifiable by transport-based
divergences~\cite{lin2023ot}---rather than an impact on the ROC,
especially since the latter is prevalence-invariant.
The two designs answer different questions in essence: Design~A asks
``how early can the renewal decision be predicted?'' In contrast,
Design~B asks ``at day~$N$, to what extent can I predict the remaining
risk?'' An illustrative comparison is \textbf{B vs.\ C}: both use 17
features as landmarks, but in C the target and cohort are fixed, while B
changes both. C \emph{rises}. So B's decline comes from the shrinking
target and re-selection, not from ``more data hurts.''

\textbf{(3) Among rising curves, the knee \emph{location} tracks the feature
set.} A and C both curve up, but the front-loading is different: A has
53\% of its ROC gain by day~45 and 89\% by day~90 (the knee); C is
\textbf{back-loaded} (34\% by day~45, 64\% by day~90, with maximal
per-day gain in $90{\to}120$). Supported by control: adding logs to
Design~C (cohort/target stays the same, only features change): logs of
Design~A reach saturation in weeks, while C's features accumulate
slowly, and \textbf{adding logs to C turns it from back-loaded to
front-loaded} (68/92\% of ROC gain by day 45/90; the $90{\to}120$ share
drops from $36\%$ to $8\%$), confirming that the feature set drives the
knee.

\subsection{The two findings the stress-test leaves standing}\label{sec:findings}
\textbf{The early indicator focuses on the contract, while the behavioral
boost becomes more evident in the high-churn-density categories.} The
profile/contract floor ($N{=}0$: auto-renew, plan, price) already
explains \textbf{70.6\% of the PR ceiling and 95.5\% of the ROC
ceiling}. The earliest indicator is one price feature
(\texttt{tx\_mean\_actual\_paid}), which has been proved not to be a
label proxy (a price-only ranker has ROC 0.77--0.81 only). The full set
of behavioral features without contract features has ROC 0.67--0.70
across windows. This is the real indicator, but it is much weaker than
the contract floor, which makes it practically useless when the contract
is known. Hence, early predictability is essentially
\textbf{contract-driven}. The extreme imbalance in the auto-renewal
category (1.55\% churn) references the class-imbalance issue described
in financial fraud
detection~\cite{sunfraud1,sunfraud2} and underlines the
appropriateness of using PR-AUC as the main measure.

\textbf{Not a mere outcome of survival.} One might think that
conditioning on $\geq$180-day survivors could artificially increase the
early ROC. Testing this premise, Design~U implements B's entire pool of
1.97M acquirees while applying a fixed churn-by-180 goal and completely
dropping the survival filter. The ROC reaches \textbf{0.913} by day~7,
and reaches 0.963 at day~120 (fold spread $\pm0.001$). The effect is
present throughout the population of acquirees. However, it should be
noted that plan span incorporates the day-180 decision into the
calculation, so this number is just a ceiling, rather than a behavioral
measure.

\emph{The contract floor consists of segments.} The composite 0.878
floor breaks down into auto-renewers at the point of acquisition
(72\%, 1.55\% churn, ROC~0.84) and \textbf{manual-renewal} users
(28\%, 25.9\% churn, \textbf{86.6\% of churners},
ROC~\textbf{0.63}), where this behavior creates a real lift (ROC
$0.64{\to}0.71$, PR~$+0.10$ by $N{=}120$).

\subsection{Cross-domain replication (OULAD)}\label{sec:oulad}
Table~\ref{tab:oulad} shows the survival-filtered OULAD curve. The curve
is \textbf{monotonically rising}, replicating the KKBox shape despite a
completely different signal regime. Three contrasts: (i)~the demographic
floor is near-chance (ROC~0.556 vs.\ KKBox's 0.878), so behavioral
features drive nearly all discrimination, the opposite of KKBox's
contract-dominated regime; (ii)~the curve is \textbf{back-loaded} (only 34\%
of PR gain by day~30; 66\% by day~90), paralleling KKBox Design~C's
back-loading when contract features are absent; (iii)~behavioral features
recover 42\% of ROC headroom (floor${\to}120$: $+0.187$ of $0.444$)
vs.\ KKBox's 19\% ($+0.023$ of $0.122$). The rising-curve shape replicates; signal composition, gain magnitude,
and curve loading do not.

\begin{table}[t]
\caption{OULAD Design-C sufficiency curve (cohort: enrolled at day~120,
$n{=}24{,}484$, churn 9.1\%; GBDT, 5-fold; fold spread
ROC $\le\pm0.027$, PR $\le\pm0.037$).}
\label{tab:oulad}
\centering\small
\begin{tabular}{rccc}
\toprule
$N$ (days) & PR-AUC & ROC-AUC & \% of PR gain \\
\midrule
0 (floor) & 0.114 & 0.556 & 0\% \\
7   & 0.116 & 0.569 & 2\% \\
14  & 0.120 & 0.587 & 5\% \\
21  & 0.136 & 0.617 & 18\% \\
30  & 0.155 & 0.635 & 34\% \\
45  & 0.173 & 0.662 & \textbf{48\%} \\
60  & 0.187 & 0.672 & 60\% \\
90  & 0.194 & 0.698 & 66\% \\
120 & 0.236 & 0.743 & 100\% \\
\bottomrule
\end{tabular}
\end{table}

\section{Discussion}\label{sec:disc}
\textbf{Model-class robustness.} The curve shape and descriptive inflection band hold
for GBDT, matched-size GBDT, and 1D-CNN (ordering
GBDT\,$>$\,CNN\,$>$\,LogReg throughout); a learned temporal representation does
not reach ``enough'' sooner. All models produce the same \emph{shape}.
This GBDT advantage is consistent with broader evidence:
Peng et~al.~\cite{peng2023} showed that GA-tuned XGBoost achieved the
highest AUC (0.991) among six classifiers in a bank-churn benchmark,
with SHAP-identified drivers echoing our contract-dominated signal.

\textbf{What is valid.} Contract terms at acquisition give the initial
clue. Behavioral lift is only important for the manual-renewal segment
(PR~$+0.10$, floor ROC~0.63), while auto-renewers account for the
greater part of cases. Design~U shows that early predictability is not
due to survivorship conditioning (ROC~0.913 at day~7 for all
acquirees); however, plan duration partly encodes the decision horizon,
making this an upper bound. In KKBox Design~A, almost 90\% of behavioral
lift attainable by day~120 can be achieved by day~90 already. This is a
descriptive observation on one dataset, not a transferable
recommendation.

\textbf{The methodological caution.} We conducted the replication study
on OULAD (\S\ref{sec:oulad}) making use of a survival filter to
eliminate the effect of in-window withdrawal leakage. The upward curve
appeared again in this behavior-dominated context; however, the signal
structure and curve loading were reversed compared to KKBox. Some
previous studies~\cite{ballings,drachen} employed a single design, which
would not detect this kind of dependence.
Multi-view learning methods~\cite{limv2024,limv2025} similarly leverage
complementary representations to improve robustness; our multi-design
stress-test applies a related principle at the experimental-design level.

\textbf{Threats to validity.} The primary design conditions on
$\geq$180-day survivors; Design~U (\S\ref{sec:findings}) checks whether
the signal holds without that filter. Label re-derivation agrees with
\texttt{train\_v2} on 92.1\% of overlap (the gap is anchor-driven;
mechanism match: 95.7\%). Because $N{=}180$ may pick up pre-decision
transactions, every headline number uses $N\le120$. We report fold means
with descriptive spread rather than paired-$t$
CIs~\cite{dietterich,nadeau}. The OULAD replication uses a Design-C
construction (landmark $L{=}120$) to avoid in-window withdrawal leakage;
it tests curve shape only, not the full design-dependence analysis.

\section{Conclusion}
When any claim is made regarding ``$N$ days shall be enough,'' what
needs to be emphasized is what kind of cohort has been constructed,
which issue is being investigated and which feature family is used in
the analysis since such a conclusion may differ depending on every
variable. The descriptive curve of KKBox has been calculated as lying
in the range of 30--120~days (Kneedle bootstrap 95\% CI: $[45,60]$),
and it can be noted that it depends solely on the analysis method,
while there are no instabilities in the data. A three-design stress
test showed that the rising trend holds when only the feature horizon
$[0,N)$ is varied, while the curve inverted under a moving-target task
and the knee location shifted with the feature set. The early signal
is delivered by contract terms, while behavioral observation is
effective only for the manual-renewal group. Design~U indicates that
early forecasting is not related to survivorship bias (i.e.\ day-7
ROC equals 0.913 for all acquirees). However, the length of the plan
has an effect on the prediction limit. A survival-filtered replication
on OULAD data confirmed the rising-curve shape in a behavior-dominated
educational setting.

\end{document}